\documentclass[letterpaper]{article} 
\usepackage[]{aaai2026}  
\usepackage{times}  
\usepackage{helvet}  
\usepackage{courier}  
\usepackage[hyphens]{url}  
\usepackage{graphicx} 
\usepackage{natbib}  
\usepackage{caption} 
\usepackage{algorithm}
\usepackage{algorithmic}

\usepackage{booktabs}     
\usepackage{multirow}     
\usepackage{amsmath}      
\usepackage{amsfonts}

\usepackage{newfloat}
\usepackage{listings}
\DeclareCaptionStyle{ruled}{labelfont=normalfont,labelsep=colon,strut=off} 
\floatstyle{ruled}
\newfloat{listing}{tb}{lst}{}
\floatname{listing}{Listing}
\title{DDFusion:Degradation-Decoupled Fusion Framework for Robust Infrared and Visible Images Fusion}
\author{
   Tianpei Zhang, Jufeng Zhao, Yiming Zhu, Guangmang Cui, Yuxin Jing
}
\usepackage{bibentry}

\begin{document}

\maketitle

\begin{abstract}
Conventional infrared and visible image fusion(IVIF) methods often assume high-quality inputs, neglecting real-world degradations such as low-light and noise, which limits their practical applicability. To address this, we propose a \textbf{D}egradation-\textbf{D}ecoupled Fusion(DDFusion) framework, which achieves degradation decoupling and jointly models degradation suppression and image fusion in a unified manner. Specifically, the Degradation-Decoupled Optimization Network(DDON) performs degradation-specific decomposition to decouple inter-degradation and degradation–information components, followed by component-specific extraction paths for effective suppression of degradation and enhancement of informative features. The Interactive Local-Global Fusion Network (ILGFN) aggregates complementary features across multi-scale pathways and alleviates performance degradation caused by the decoupling between degradation optimization and image fusion. Extensive experiments demonstrate that DDFusion achieves superior fusion performance under both clean and degraded conditions.  Our code is available at https://github.com/Lmmh058/DDFusion.
\end{abstract}

\section{Introduction}
Infrared and visible light images exhibit strong complementary characteristics. Specifically, visible images have rich texture details but are susceptible to variations in lighting, while infrared images provide thermal information but lack texture details. Therefore, IVIF can overcome the limitations of noise and low resolution in infrared images, as well as the effects of lighting and camouflage on visible images. This capability is crucial in applications such as autonomous driving \cite{abrecht2024deep} and night-time target detection \cite{hu2025datransnet, xiao2024background}. 

In real-world scenarios, source images are often degraded by sensor limitations or environmental factors, such as low-light conditions in visible images and noise in infrared ones. Direct fusion of such inputs may introduce artifacts and information loss, leading to suboptimal IVIF performance. Hence, IVIF methods tailored to degraded inputs are crucial for practical deployment.

\begin{figure}[!t]
    \centering
    \includegraphics[width=0.85\columnwidth]{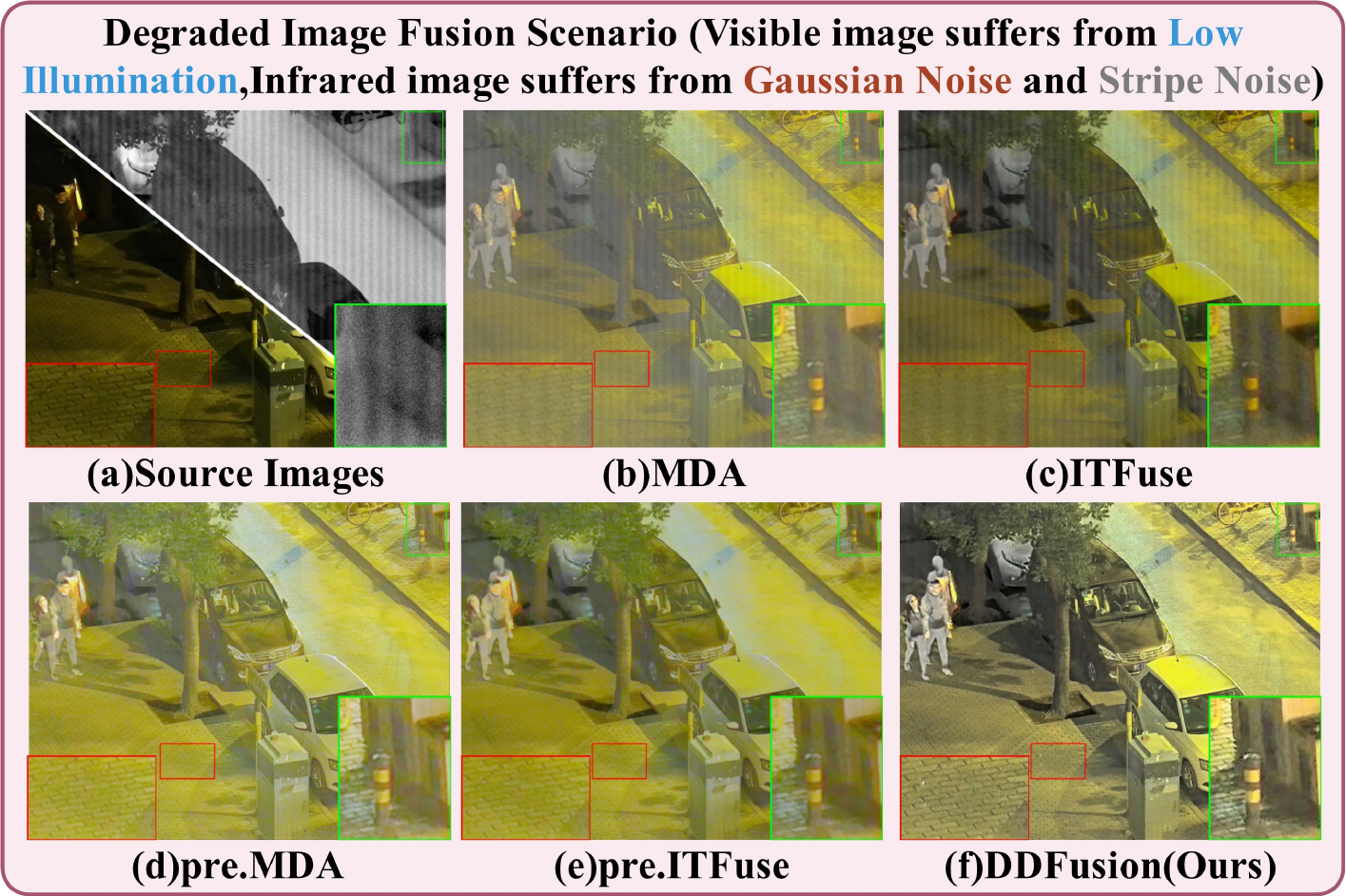}    \caption{Compound degradation image fusion. \textit{pre.} denotes degradation-specific pre-processing.}
    \label{fig:figure1}
\end{figure}

Deep learning-based IVIF methods have evolved into diverse frameworks \cite{liu2021learning,ma2022swinfusion}, demonstrating significant progress under high-quality image assumptions. However, these approaches largely overlook the prevalent sensor-induced degradations in real-world scenarios, such as luminance distortion in visible images \cite{tang2022piafusion} and Gaussian \cite{chen2021Gaussian} or stripe noise \cite{wang2016stripe} in infrared images. When source images suffer from compound degradation, existing methods face dual challenges: (1) Fusion networks built on ideal imaging assumptions tend to inherit degradation artifacts (Fig.\ref{fig:figure1}(b,c)). (2) Although some approaches tackle single degradation types via dedicated networks or adopt two-stage pipelines combining preprocessing and fusion, they remain limited in handling multi-modal compound degradation due to information loss from preprocessing and the decoupling of optimization and fusion (Fig.\ref{fig:figure1}(d,e)).

\begin{figure}
    \centering
    \includegraphics[width=0.85\columnwidth]{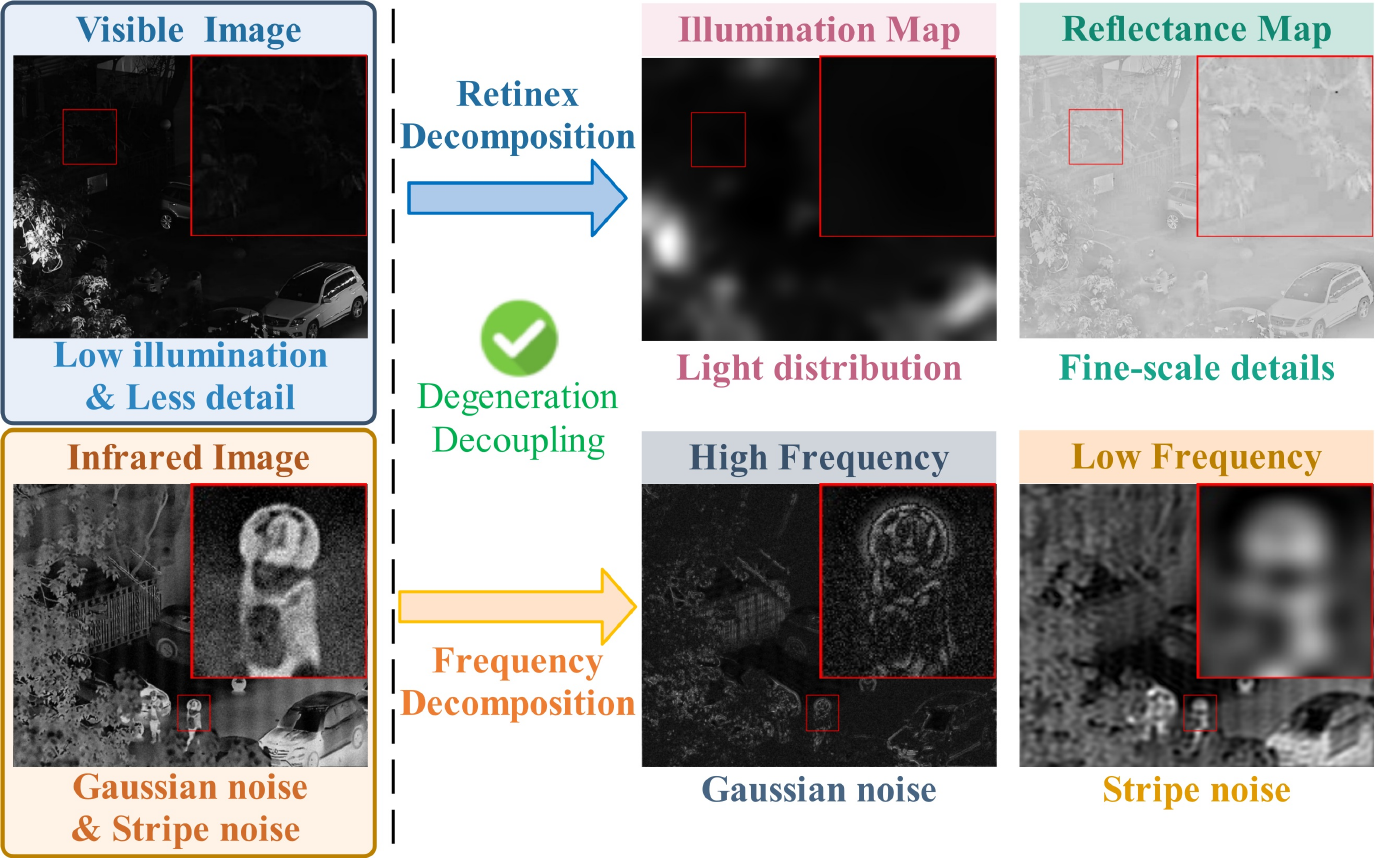}
    \caption{Degraded infrared and visible images are decomposed via frequency and Retinex methods: visible images yield separated detail and degraded luminance, while infrared images reveal Gaussian and stripe noise.}
    \label{fig:fig2}
\end{figure}

To address the limitations, we argue that effective fusion of images with compound degradations hinges on two key aspects: accurate degradation separation and the joint optimization of degradation suppression and image fusion. Among them, degradation separation plays a pivotal role, enabling targeted adjustment and suppression based on the distinct characteristics of each degradation type. As shown in Fig.\ref{fig:fig2}, illumination degradation in visible images can be isolated via Retinex-based decomposition, while compound noise in infrared images can be disentangled through frequency-domain analysis. Such decomposition facilitates the separation of complex degradations, laying a foundation for component-specific degradation modeling and information preservation.

Based on the above analysis, we propose the DDFusion framework, comprising two core components: DDON and ILGFN. To achieve accurate degradation separation, DDON performs degradation decoupling for infrared and visible images via 2D-DCT \cite{ahmed2006discrete} and Retinex \cite{land1971retinex}, respectively, and applies targeted feature extraction strategies to jointly suppress degradation and extract useful information from the decomposed components. To enable joint optimization of suppression and feature fusion, ILGFN integrates the enhanced features from both modalities output by DDON via multi-scale local and global interactive attention, effectively aggregating complementary information and mitigating the decoupling between degradation modeling and image fusion. The main contributions are as follows:

\begin{enumerate}
    \item We propose DDFusion, which enables joint modeling of compound degradation decoupling, optimization, and fusion, effectively reducing the impact of composite degradations on the fusion results.
    \item We propose DDON, which performs modality-specific degradation decoupling and couples degradation optimization with feature extraction.
    \item We propose ILGFN, which aggregates features across multi-scale pathways and mitigates the decoupling between degradation optimization and feature fusion.
\end{enumerate}

\section{Related Work} \label{sec:related}

\subsection{Infrared and Visible Image Fusion}


The flourishing development of deep learning methods has revitalized new vitality into the IVIF, with diverse frameworks emerging: Autoencoders(AE) \cite{li2021rfn, liu2021learning}, convolutional neural networks(CNN) \cite{xu2020u2fusion, tang2022piafusion}, generative adversarial networks(GAN) \cite{ma2020ganmcc, ma2020ddcgan}, and vision transformer(ViT) \cite{zhu2024towards, ma2022swinfusion}. Additionally, several approaches \cite{lv2024sigfusion, liu2022target} aim to retain richer semantic information to enhance the performance of image fusion in downstream tasks. However, these methods are designed for high-quality image fusion and tend to retain significant degradation artifacts when applied to real-world degraded inputs.

\begin{figure*}[!t]
    \centering
    \includegraphics[width=0.9\textwidth]{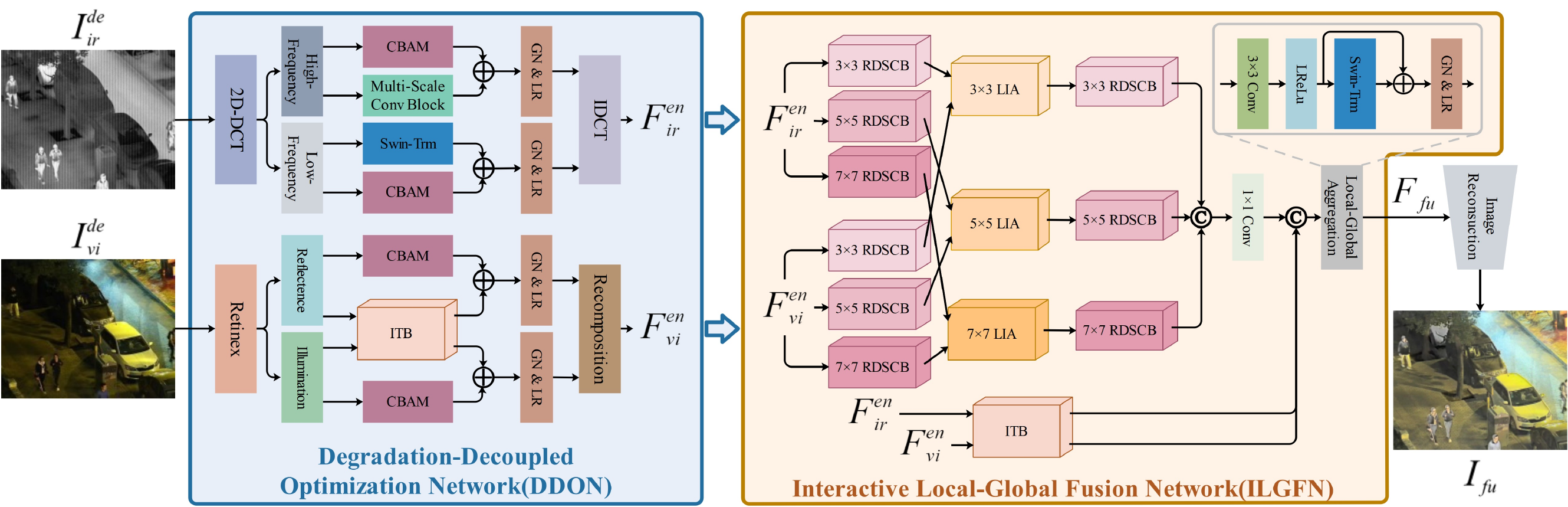}
    \caption{Overall architecture of the proposed DDFusion. The DDFusion framework comprises two principal subnetworks: the Degradation-Decoupled Optimization Network (DDON) and the Interactive Local-Global Fusion Network (ILGFN).}
    \label{fig:Network}
\end{figure*}

\begin{figure}[!t]
    \centering
    \includegraphics[width=0.8\columnwidth]{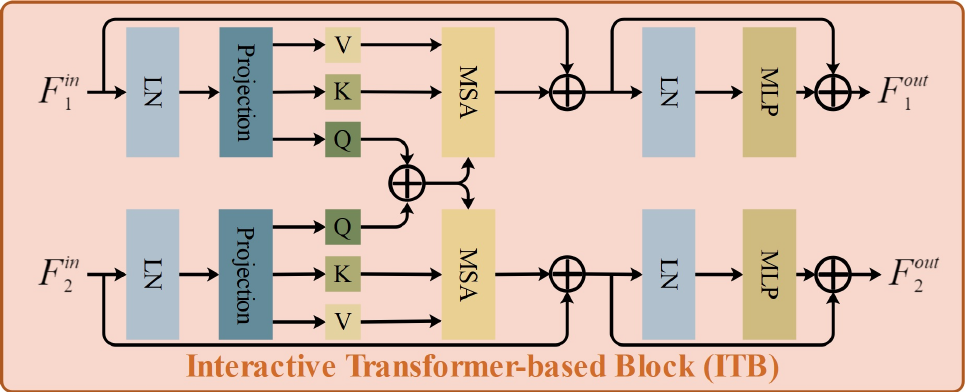}
    \caption{Architecture of Interactive Transformer-based Block(ITB)}
    \label{fig:ITB}
\end{figure}

\subsection{Degradation-Aware image fusion}
To address the degradation-induced quality drop in fused images, some studies have designed specialized fusion networks for single-type degradation \cite{yang2024iaifnet,tang2023divfusion} tailored for specific issues like low-light visible images or Gaussian-noisy infrared images but fail to generalize to compound degradations. Text-guided modulation methods \cite{yi2024text} introduce semantic prompts to target degradation removal but when dealing with compound degradation—e.g., luminance distortion and mixed noise  still rely on additional preprocessing and suffer from decoupled degradation optimization and fusion, leading to inevitable information loss.

\section{Method} \label{sec:method}
\subsection{Overall Framework}
Network architecture of DDFusion is illustrated in Fig.\ref{fig:Network}, we first convert the visible image to YCbCr color space and extract the Y component with luminance degradation as the visible input. This choice is motivated by YCbCr's unique property where the Y channel independently encodes luminance information. The degraded inputs from both modalities \(I_{vi}^{de},I_{ir}^{de}\) are then processed through DDON to achieve degradation suppression and effective feature mining:
{\footnotesize
\begin{equation}
\label{eq:DDON}
F_{ir}^{en},F_{vi}^{en} = DDON(I_{ir}^{de},I_{vi}^{de})
\end{equation}
}
where $F_{vi}^{en}$ and $F_{ir}^{en}$ denote the degradation-suppressed and modality-specific features, which are subsequently processed by $ILGFN$ for deep feature mining and cross-modal complementary aggregation across multi-scale pathways:
{\footnotesize
\begin{equation}
\label{eq:ILGFN}
F_{fu} = ILGFN(F_{vi}^{en},F_{ir}^{en})
\end{equation}}
where, $F_{fu}$ denotes the fused features. Finally, an image reconstruction module $Re$, consisting of three consecutive 3×3 convolutions with LeakyReLU activations, maps $F_{fu}$ to the image space to produce the output $I^Y_{fu} = Re(F_{fu})$. This output serves as the new Y channel and is recombined with the Cb and Cr components from the visible image, followed by conversion to RGB to obtain the final fused result $I_{fu}$.

\subsection{Interactive Transformer-based Block}
Although Swin-Transformer \cite{liu2021swin} offers strong global feature extraction, it struggles to model the global complementarity between different input modulations in DDFusion. To address this limitation, we propose the Interactive Transformer-based Block (ITB), as shown in Fig.\ref{fig:ITB}, to enable cross-component interaction between input features $F^{in}_{1}$ and $F^{in}_{2}$:
{\footnotesize
\begin{equation}
\label{eq:ITB}
\begin{aligned}
&F^{ISA}_{1} = ISA(LN(F^{in}_{1}),LN(F^{in}_{2})) + F^{in}_{1} \\
&F^{out}_{1} = MLP(LN(F^{ISA}_{1})) + F^{ISA}_{1} \\
&F^{ISA}_{2} = ISA(LN(F^{in}_{2}),LN(F^{in}_{1})) + F^{in}_{2} \\
&F^{out}_{2} = MLP(LN(f^{ISA}_{2})) + F^{ISA}_{2} \\
\end{aligned}
\end{equation}}
where $F^{out}_{1}, F^{out}_{2}$ represent the dual outputs of the ITB. Unlike multi-head self-attention(MSA), our interactive self-attention(\(ISA\)) mechanism enables cross-component interaction through shared-query computation. For input windows \(X^{in}_{1}, X^{in}_{2} \in \mathbb{R}^{M^2 \times C}\), the ISA operates as:  
{\footnotesize
\begin{equation}
\label{eq:ISA}
\begin{aligned}
&\{Q_{1},K_{1},V_{1}\} = \{X^{in}_{1}W_{Q1},X^{in}_{1}W_{K1},X^{in}_{1}W_{V1}\} \\
&\{Q_{2},K_{2},V_{2}\} = \{X^{in}_{2}W_{Q2},X^{in}_{2}W_{K2},X^{in}_{2}W_{V2}\} \\
&X^{out}_{1} = MSA(Q_{1}+Q_{2},K_{1},V_{1}) \\
&X^{out}_{2} = MSA(Q_{2}+Q_{1},K_{2},V_{2})
\end{aligned}
\end{equation}}
where \(X^{out}_1, X^{out}_2\in\mathbb{R}^{M^2 \times C}\) denote the output windows.

\subsection{Degradation-Decoupled Optimization Network}
The architecture of the Degradation-Decoupled Optimization Network (DDON) is shown in Fig.\ref{fig:Network}. Specifically, 2D-DCT decomposition ($2DDCT(\cdot)$) is applied to the degraded infrared image $I^{de}_{ir}$ to separate high-frequency components $F_{high}$ dominated by Gaussian noise and low-frequency components $F_{low}$ containing stripe noise. In parallel, Retinex decomposition ($Retinex(\cdot)$) is applied to the degraded visible image $I^{de}_{vi}$ to decouple the degraded illumination component $F_l$ and the detail-rich reflectance component $F_r$. This process can be formulated as:
{\footnotesize
\begin{equation}
\label{eq:decomposition}
\begin{aligned}
   &\{F_{low},F_{high}\} = 2DDCT(I^{de}_{ir})\\
   &\{F_{r},F_{l}\} = Retinex(I^{de}_{vi}) 
\end{aligned}
\end{equation}}
subsequently, we design component-specific feature extraction paths to enable effective degradation suppression and feature mining: for $F_{low}$, a multi-scale convolution block $MSConv(\cdot)$ with kernel sizes of 1, 3, 5, and 7 is used to extract detail features and filter Gaussian noise under diverse receptive fields; a SwinTransformer block $SwinTrm(\cdot)$ is applied to model structural features and suppress stripe noise in $F_{low}$; for $F_{r}$ and $F_{l}$, an Interactive Transformer Block $ITB(\cdot)$ is used to enhance reflectance details and adaptively adjust illumination distribution. In addition, each component is refined with CBAM attention \cite{woo2018cbam} to emphasize critical information. This process is formulated as:
{\footnotesize
\begin{equation}
\label{eq:enhancement}
\begin{aligned}
&F'_{high} = MSConv(F_{high})\\
&F'_{low} = SwinTrm(F_{low}) \\
&\{F'_{r},F'_{l}\} = ITB(F_{r},F_{l}) \\
&F_{x}^{en} = GN\&LR(F'_{x} + CBAM(F_{x}))
\end{aligned}
\end{equation}}
where $x\in\{high,low,r,l\}$, and $F_{x}^{en}$ denotes the enhanced components. $GN \& LR$ represents group normalization with LeakyReLU. The enhanced infrared feature is reconstructed via inverse DCT as $F_{ir}^{en} = IDCT(F_{low}^{en}, F_{high}^{en})$, and the enhanced visible feature is computed via element-wise multiplication: $F_{vi}^{en} = F_{r}^{en} \otimes F_{l}^{en}$.

\subsection{Interactive Local-Global Fusion Network}
To enable effective cross-modal feature interaction and alleviate performance degradation caused by the decoupling of degradation optimization and fusion, we design the ILGFN, as shown in Fig.\ref{fig:Network}. Fusion is performed across three local paths with kernel sizes of 3, 5, and 7, and one global path, using the enhanced modality features from DDON. In the local paths with kernel size $n$, $F^{en}_{ir}$ and $F^{en}_{vi}$ are processed by multi-scale Residual Depthwise Separable Convolution Blocks ($RDSCB_{ir}^{n}, RDSCB_{vi}^{n}$) for deep local feature extraction. Features of the same scale are then fused via corresponding Local Interaction Attention($LIA$) and $RDSCB^{n}$ modules to highlight and aggregate complementary features. In the global path, $F^{en}_{ir}$ and $F^{en}_{vi}$ are complementarily enhanced through global feature interaction via $ITB$. The overall process is defined as:
{\footnotesize
\begin{equation}
\label{eq:fusion}
\begin{aligned}
&F_{vi}^{n} = RDSCB_{vi}^{n}(F_{vi}^{en}), F_{ir}^{n} = RDSCB_{ir}^{n}(F_{vi}^{en})\\
&F_{fu}^{n} = RDSCB^{n}(LIA(F_{vi}^{n},F_{ir}^{n})) \\ 
&F^{l}_{fu} = Conv_{1}(Concat(F_{fu}^{n})),n\in \{3,5,7\} \\
&\{F_{vi}^{g},F_{ir}^{g}\} = ITB(F_{vi}^{en},F_{ir}^{en}))
\end{aligned}
\end{equation}}
where $Concat$ denotes channel-wise concatenation and $Conv_{1}$ indicates a 1×1 convolution. $F^{l}_{fu}, F^{g}_{vi},$ and $F^{g}_{ir}$ represent the fused local features and the global features from the visible and infrared modalities, respectively. Finally, the fused feature $F_{fu}$ is obtained via Local-Global Aggregation(LGA), which consists of sequential convolutional operations and SwinTransformer blocks:

{\footnotesize
\begin{equation}
\label{eq:fusion}
F_{fu} = LGA(Concat(F^{l}_{fu},F_{vi}^{g},F_{ir}^{g}))
\end{equation}}

\begin{figure}[!t]
    \centering
    \includegraphics[width=0.75\columnwidth]{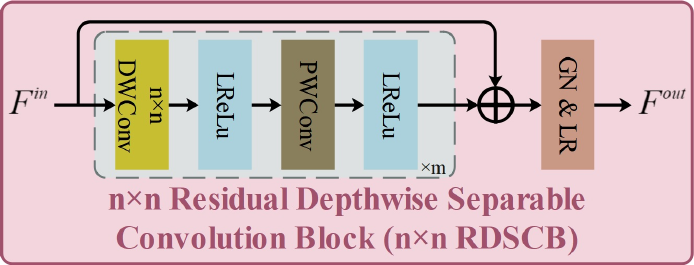}
    \caption{Architecture of Residual Depthwise Separable Convolution Block(RDSCB), where $n$ denotes the depthwise convolution kernel size and $m$ indicates the repetition count of the module within the gray block.}
    \label{fig:RDSCB}
\end{figure}

\subsubsection{Residual Depthwise Separable Convolution Block:}
The architecture of RDSCB (Fig.\ref{fig:RDSCB}) integrates depthwise convolution with kernel size $n$ for scale-specific spatial aggregation and pointwise convolution for channel interaction. This combination is repeated $m$ times to enrich feature representations. Given input $F^{in}$, the process is defined as:
{\footnotesize
\begin{equation}
\label{eq:RDSCB}
\begin{aligned}
&F = (LR(PWConv(LR(DWConv_{n}(F^{in})))))^m \\ 
&F^{out} = GN \& LR(F+F^{in}))
\end{aligned}
\end{equation}}

\subsubsection{Local Interaction Attention:}
\begin{figure}[!t]
    \centering
    \includegraphics[width=0.8\columnwidth]{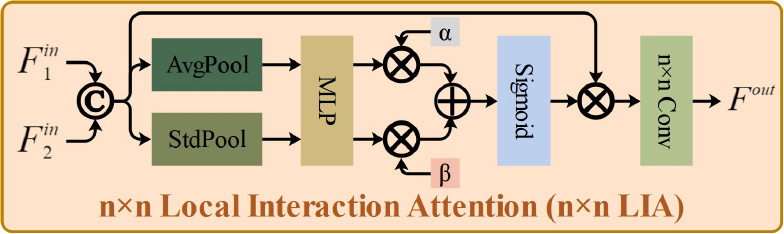}
    \caption{Architecture of Local Interaction Attention(LIA), with $n$ denoting the convolution kernel size.}
    \label{fig:LIA}
\end{figure}

To fully exploit local cross-modal complementarity, we propose the LIA mechanism (Fig.\ref{fig:LIA}). Given input features $F^{in}_1$ and $F^{in}_2$, spatial average pooling ($Avg$) and standard deviation pooling ($Std$) are used to extract modality saliency, followed by an MLP for nonlinear channel dependency modeling. Learnable parameters $\alpha$ and $\beta$ enable adaptive weighting, while a Sigmoid function produces attention maps to modulate the original features. Finally, a convolution with kernel size $n$ is applied for feature reconstruction. The process is defined as:
{\footnotesize
\begin{equation}
\label{eq:LIA}
\begin{aligned}
&F' = Concat(F^{in}_{1},F^{in}_{2}) \\ 
&Att = \alpha MLP(Avg(F')) + \beta MLP(Std(F')) \\
&F^{out} = Conv_{n}(Sigmoid(Att) \otimes F')
\end{aligned}
\end{equation}}

\subsection{Loss Function}
\subsubsection{Degradation Optimization Loss:}
To ensure that DDON effectively suppresses degradation while preserving useful information, we design a degradation optimization loss function composed of three components:
{\footnotesize
\begin{equation}
\label{eq:degrade loss}
\mathcal{L}_{do} = \mathcal{L}_{ir} + \mathcal{L}_{vi} + \mathcal{L}_{ds}
\end{equation}}
where $\mathcal{L}_{ir}$ and $\mathcal{L}_{vi}$ denote the reconstruction losses for the infrared and visible modalities, respectively, aiming to preserve pixel-level informative content. The reconstruction loss is defined as:
{\footnotesize
\begin{equation}
\label{eq:irvi loss}
\mathcal{L}_{y} = \frac{1}{HW} \sqrt{(I^{en}_{y} - I^{ref}_{y})^2 + \varepsilon}, y \in \{ir, vi\}
\end{equation}}
where $I^{ref}_{y}$ is the reference image and $I^{en}_{y}$ is the reconstructed image for modality $y$, and $\varepsilon$ is a small constant set to $1 \times 10^{-6}$. In addition, we introduce a degradation suppression loss $\mathcal{L}_{ds}$, which comprises illumination loss $\mathcal{L}_{illu}$, total variation loss $\mathcal{L}_{tv}$, and infrared perceptual loss $\mathcal{L}_{per}$:
{\footnotesize
\begin{equation}
\label{eq:reconstrction loss}
\mathcal{L}_{ds} = \lambda_{1}\mathcal{L}_{illu} + \lambda_{2}\mathcal{L}_{tv} +\lambda_{3} \mathcal{L}_{per}
\end{equation}}
where $\lambda_{1}, \lambda_{2}, \lambda_{3}$ are weighting factors. The illumination loss $\mathcal{L}_{illu}$ encourages a more natural and smooth luminance distribution in the visible image by minimizing the average brightness difference across patches, and is defined as:
{\footnotesize
\begin{equation}
\label{eq:illumination_loss}
\begin{aligned}
&L^{en}_{vi} = Avg_{16}(I^{en}_{vi}) ,L^{ref}_{vi} = Avg_{16}(I^{ref}_{vi})\\
&\mathcal{L}_{illu} = \frac{16 \times 16}{HW} \sqrt{(L^{en}_{vi} - L^{ref}_{vi})^2 + \varepsilon}
\end{aligned}
\end{equation}}
where, $Avg_{16}$ denotes average pooling with a patch size of 16. The total variation loss $\mathcal{L}_{tv}$ smooths the image by minimizing its total variation to reduce noise:
{\footnotesize
\begin{equation}
\label{eq:tv_loss}
\begin{aligned}
    \mathcal{L}_{tv}(x) = \frac{1}{HW} \sum_{i,j} \left[(x_{i+1,j} - x_{i,j})^2 + (x_{i,j+1} - x_{i,j})^2 \right]
\end{aligned}
\end{equation}}
where $x_{i,j}$ denotes the infrared pixel value in $I^{en}_{vi}$ at $(i,j)$. Finally, we apply a perceptual loss $\mathcal{L}_{per}$ to reinforce feature inheritance from the source images.
{\footnotesize
\begin{equation}
\label{eq:lossint}
  \mathcal{L}_{per} = ||VGG(I^{en}_{ir})-VGG(I^{ref}_{ir})||^{2}_{2}
\end{equation}}
where \( VGG(\cdot) \) denotes the pre-trained VGG-16 model.

\begin{figure}[!t]
\centering
\includegraphics[width=\columnwidth]{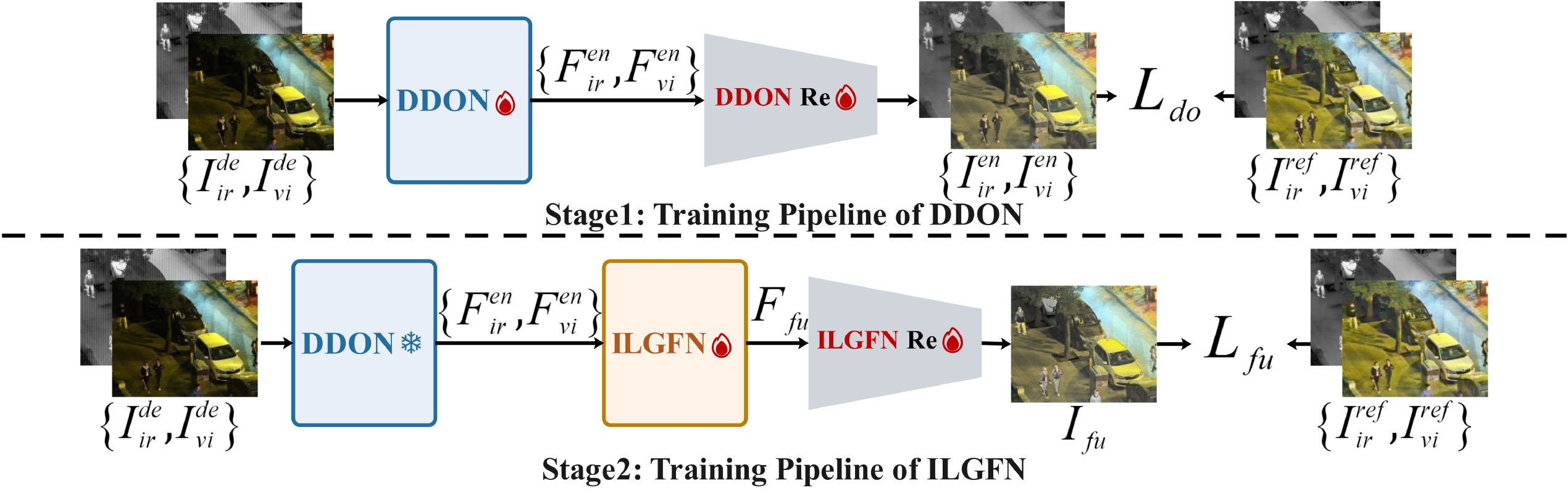}
\caption{Training pipeline of DDFusion. DDON parameters are frozen during ILGFN training.}
\label{fig:traning_pipeline}
\end{figure}

\begin{table}[!t]
\centering
{\small
\begin{tabular}{ll}
\toprule[1pt]
Methods(\textit{Source \& Year \& Type})     \\ \hline
GANMcC \cite{ma2020ganmcc} \textit{TIM'2020, GAN-based} \\ 
RFN-Nest \cite{li2021rfn} \textit{INFFUS'2021, AE-based} \\ 
CUFD \cite{xu2022cufd} \textit{CVIU'2022, AE-based}\\
DATFuse \cite{tang2023datfuse} \textit{TCSVT'2023, Transformer-based}  \\ 
Fusionmamba \cite{xie2024fusionmamba} \textit{2024, Mamba-based} \\
MDA \cite{yang2024mda} \textit{NEUCOM'2024, CNN-based} \\
ITFuse \cite{tang2024itfuse} \textit{PR'2024, Transformer-based}  \\
\hline
$\star$ \textbf{DDFusion(Ours), Transformer-based} \\
\bottomrule[1pt]
\end{tabular}}
\caption{Configurations for all comparative experiments}
\label{tab:comparative methods}
\end{table}

\begin{figure*}[!t]
\centering
\includegraphics[width=0.9 \textwidth]{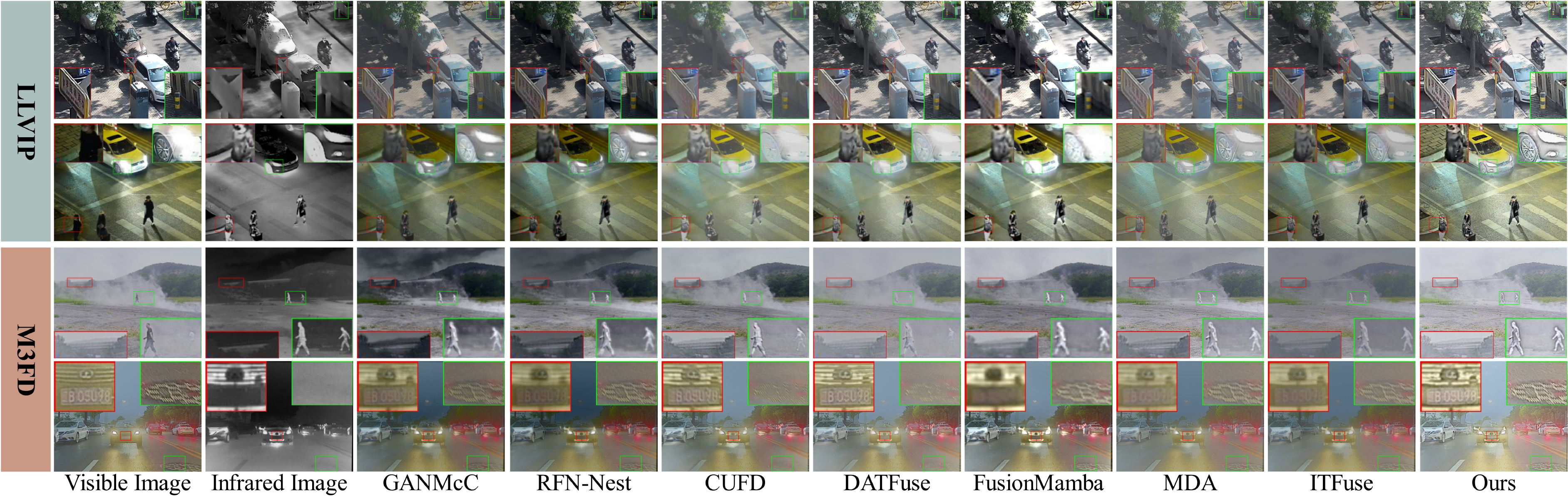}
\caption{Qualitative comparison of DDFusion under non-degraded conditions on the LLVIP and M3FD datasets, key regions are enlarged to simplify visual comparison.}
\label{Qualitative}
\end{figure*}

\begin{table*}[!t]
\centering
{\small
\begin{tabular}{c|cccccc}
\toprule
\multirow{2}{*}{\textbf{Method}} 
& \multicolumn{6}{c}{\textbf{Image Fusion under Non-degraded Conditions(LLVIP/M3FD)}} 
\\
\cline{2-7}
& \textbf{VIF$\uparrow$} & \textbf{AG$\uparrow$} & \textbf{EI$\uparrow$} & \textbf{Q$_{abf}\uparrow$} & \textbf{SF$\uparrow$} & \textbf{Q$_{w}\uparrow$}  \\
\midrule
GANMcC
& 0.568/0.591 & 2.144/2.396 & 22.304/25.116 & 0.260/0.341 & 7.070/6.967 & 0.482/0.589
\\
RFN-Nest  
& 0.633/0.626 & 2.234/2.450 & 23.777/26.205 & 0.312/0.393 & 6.832/6.893 & 0.489/0.614
\\
CUFD    
& 0.570/0.635 & 2.294/3.212 & 24.122/33.721 & 0.323/0.391 & 8.116/9.422 & 0.520/0.568
\\
DATFuse   
& 0.701/0.709 & 2.967/3.030 & 30.514/30.827 & \textbf{0.470}/0.494 & 11.321/9.478 & 0.684/0.588 \\
FusionMamba 
& 0.558/0.462 & 1.988/2.455 & 21.963/26.856 & 0.203/0.221 & 5.712/6.635 & 0.213/0.338
\\
MDA
& 0.494/0.502 & 2.054/2.739 & 21.256/28.464 & 0.189/0.330 & 6.822/8.425 & 0.377/0.556
\\
ITFuse
& 0.566/0.489 & 1.839/1.729 & 19.636/18.427 & 0.212/0.204 & 5.600/4.778 & 0.386/0.377
\\
\textbf{$\star$Ours}     
& \textbf{1.005}/\textbf{0.796} & \textbf{6.332}/\textbf{4.597} & \textbf{64.619}/\textbf{47.262} & 0.464/\textbf{0.561} & \textbf{19.832}/\textbf{14.217} & \textbf{0.740}/\textbf{0.804}\\
\bottomrule
\end{tabular}
}
\caption{Quantitative comparison of DDFusion under non-degraded conditions on the LLVIP and M3FD datasets. The best results for each metric are bolded, and $\uparrow$ indicates that higher values denote better performance.}
\label{tab:Quantitative_nondegrade}
\end{table*}

\subsubsection{Fusion Loss:}
To ensure the fused image preserves adequate intensity information and detailed features from the source images, following loss components:
{\footnotesize
\begin{equation}
\label{eq:fusionloss}
  \mathcal{L}_{fu} = \gamma_{1}\mathcal{L}_{int} + \gamma_{2}\mathcal{L}_{text}
\end{equation}}
where \(\gamma_{1}\) and \(\gamma_{2}\) are weighting coefficients. We employ the intensity loss function \(\mathcal{L}_{int}\) to ensure sufficient intensity information preservation in the fused image:
{\footnotesize
\begin{equation}
\label{eq:lossint}
  \mathcal{L}_{int} = \frac{1}{HW}(||I_{fu}-I^{ref}_{ir}||_{1} +||I_{fu}-I^{ref}_{vi}||_{1})
\end{equation}}

Additionally, we introduce the texture loss \(\mathcal{L}_{text}\) to maximally preserve textures derived from the source images:
{\footnotesize
\begin{equation}
  \mathcal{L}_{text} = \frac{1}{HW}|||\nabla I_{fu}|-max(|\nabla I^{ref}_{ir}|, |\nabla I^{ref}_{vi}|)||_{1}
  \label{eq:lossgrad}
\end{equation}}
where $\nabla$ is Sobel gradient operator, $max(\cdot)$ denotes the element-wise maximum selection.

\section{Experiments} \label{sec:experiment}
\subsection{Experimental Settings} \label{subsec:setting}


\begin{figure*}[!t]
\centering
\includegraphics[width=0.9\textwidth]{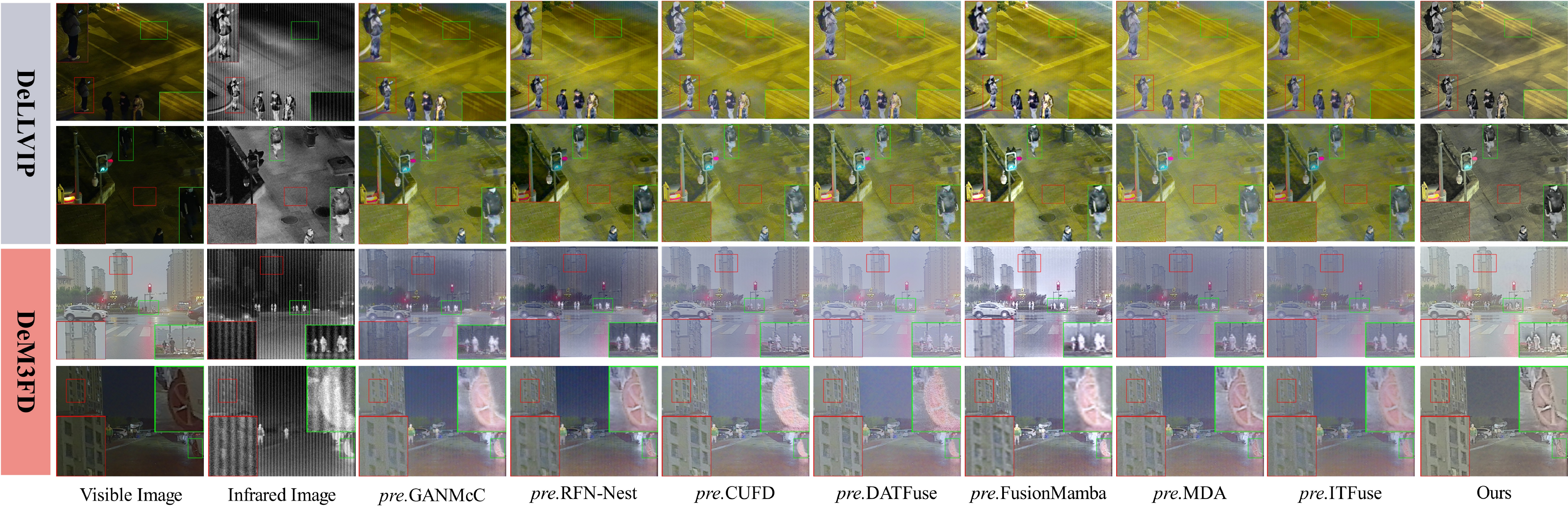}
\caption{Qualitative comparison of DDFusion under significant degradation on the DeLLVIP and DeM3FD datasets, where the visible image suffers from low-light degradation and the infrared image is affected by Gaussian and/or stripe noise.
key regions are enlarged to simplify visual comparison."pre." denotes various preprocessing methods (Zero-DCE++: low-light enhancement, ASCNet: stripe noise removal, Swin-IR: Gaussian noise removal)}
\label{DegenerateQualitative}
\end{figure*}

\begin{table*}[!t]
\centering
{\small
\begin{tabular}{c|cccccc}
\toprule
\multirow{2}{*}{\textbf{Method}} 
& \multicolumn{6}{c}{\textbf{Image Fusion under Significant Degradation(DeLLVIP/DeM3FD)}} \\
\cline{2-7} 
& \textbf{VIF$\uparrow$} & \textbf{AG$\uparrow$} & \textbf{EI$\uparrow$} & \textbf{Q$_{abf}\uparrow$} & \textbf{SF$\uparrow$} & \textbf{Q$_{w}\uparrow$} \\
\midrule
\textit{pre.}GANMcC
& 0.645/0.446 & 3.751/2.797 & 37.982/28.063 & 0.180/0.234 & 11.821/8.319 & 0.347/0.455 \\
\textit{pre.}RFN-Nest  
& 0.764/0.477 & 3.753/2.494 & 39.625/26.516 & 0.200/0.244 & 11.077/7.253 & 0.346/0.437 \\
\textit{pre.}CUFD    
& 0.678/0.548 & 4.795/3.703 & 49.194/38.324 & 0.218/0.282 & 14.522/10.808 & 0.426/0.477 \\
\textit{pre.}DATFuse   
& 0.861/0.642 & 6.239/3.225 & 60.079/32.210 & 0.293/0.345 & \textbf{19.891}/10.282 & 0.507/0.479 \\
\textit{pre.}FusionMamba 
& 0.488/0.330 & 3.509/3.064 & 37.708/33.376 & 0.125/0.164 & 9.387/8.025 & 0.157/0.280 \\
\textit{pre.}MDA
& 0.702/0.424 & 3.911/3.007 & 40.795/31.107 & 0.163/0.229 & 11.445/9.285 & 0.305/0.432 \\
\textit{pre.}ITFuse
& 0.654/0.356 & 3.444/1.832 & 35.811/18.914 & 0.156/0.139 & 10.339/5.304 & 0.297/0.278 \\
\textbf{$\star$Ours}     
& \textbf{0.919}/\textbf{0.756} & \textbf{6.326}/\textbf{4.540} & \textbf{64.527}/\textbf{46.602} & \textbf{0.328}/\textbf{0.399} & 19.793/\textbf{14.102} & \textbf{0.584}/\textbf{0.626} \\
\bottomrule
\end{tabular}
}
\caption{Quantitative comparison of DDFusion under significant degradation on the DeLLVIP and DeM3FD datasets. The best results for each metric are bolded, and $\uparrow$ indicates that higher values denote better performance.}
\label{tab:Quantitative_degrade}
\end{table*}

\subsubsection{Implementation details:} 
\label{DeLLVIP}
The training pipeline of DDFusion is illustrated in Fig.\ref{fig:traning_pipeline}. Since existing IVIF datasets are designed for high-quality image fusion and lack sufficient degraded image pairs, we construct a training set by randomly selecting 4,725 pairs from the LLVIP dataset \cite{jia2021llvip}—some of which already contain low-light visible images—and synthetically adding Gaussian noise with mean 0 and $\sigma \in [5,30]$ and stripe noise with intensity in [10, 30] to infrared images. Training images are randomly cropped into 128$\times$128 patches and normalized before being fed into the network. The learning rate and batch size are set to 1e-3 and 16, respectively. The reference visible images used for loss computation are generated by enhancing the source visible images with Zero-DCE++ \cite{li2021zerodce++}, while the infrared references are the clean images before noise injection. The weights for $\mathcal{L}_{do}$ are set to 1, 100, and 1; for $\mathcal{L}_{fu}$, the weights are 1 and 5. All experiments are conducted on an NVIDIA GeForce RTX 4090D GPU.

We perform image fusion on 50 and 49 image pairs from the LLVIP and M3FD datasets, respectively, under conditions where source images have no significant degradation. To evaluate performance under real-world degradation, we simulate two degraded test sets—DeLLVIP and DeM3FD—by randomly selecting another 50 and 49 pairs from LLVIP and M3FD and applying the same degradation simulation used for the training set.
\subsubsection{Comparative methods and metrics:}
The comparative methods are listed in Tab.\ref{tab:comparative methods}. Six metrics are employed for quantitative evaluation: visual information fidelity(VIF) \cite{han2013new}, average gradient(AG) \cite{cui2015detail}, edge intensity(EI) \cite{rajalingam2018hybrid}, gradient-based similarity measure($Q_{abf}$) \cite{xydeas2000objective}, spatial frequency(SF) \cite{eskicioglu1995image}, and Piella’s metric($Q_{w}$) \cite{piella2003new}.

\subsection{Image Fusion under Non-degraded Conditions}
\subsubsection{Qualitative Analysis:}
Fig.\ref{Qualitative} presents the qualitative comparison of DDFusion under non-degraded conditions. GANMcC and RFN-Nest tend to inherit background intensity from infrared images, resulting in low contrast (red and green boxes in rows 3 and 4). CUFD and DATFuse struggle to preserve visible details and salient infrared targets, leading to information loss (red box in row 1 and green box in row 3). FusionMamba fails to extract sufficient details from both modalities, causing blurred or missing structures (green boxes in rows 1 and 2). MDA and ITFuse suffer from low contrast in their fused results (green box in row 1 and red box in row 2). In contrast, DDFusion produces fused images with rich detail and balanced intensity.

\subsubsection{Quantitative Analysis:}
Tab.\ref{tab:Quantitative_nondegrade} presents the quantitative results of DDFusion under non-degraded conditions. Except for a slight drop in $Q_{abf}$ on LLVIP compared to DATFuse, DDFusion outperforms all others across the remaining metrics, indicating superior feature preservation, visual fidelity, and structural consistency.

\begin{figure*}[!t]
\centering
\includegraphics[width= 0.8\textwidth]{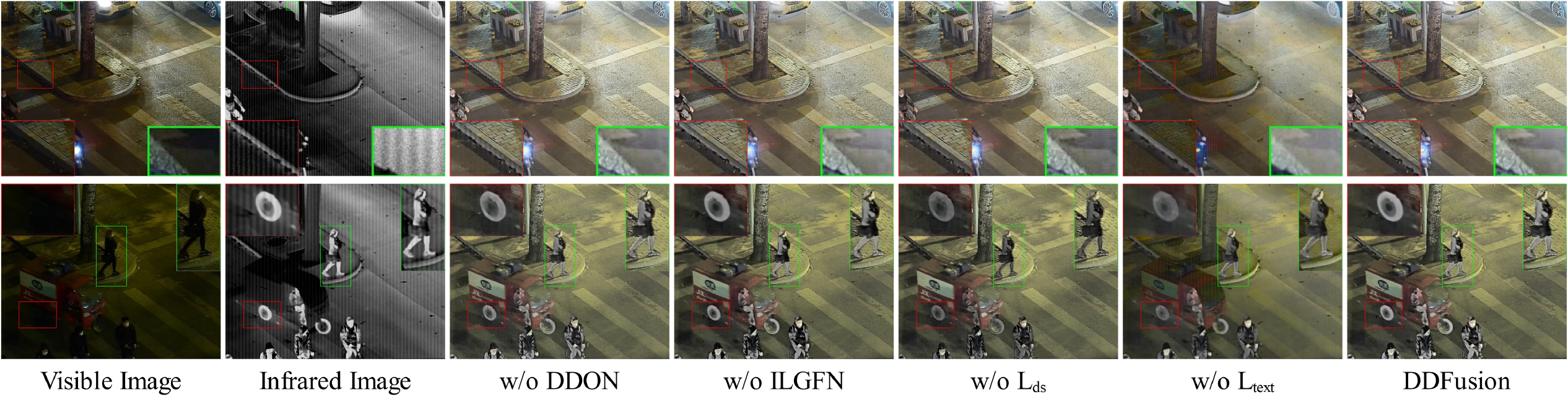}
\caption{Qualitative comparison of DDFusion ablation studies, with key regions magnified for visual inspection.}
\label{Ablation}
\end{figure*}

\subsection{Image Fusion under Significant Degradation}
To demonstrate DDFusion's superior performance under compound degradations, we evaluate it against other methods on DeLLVIP and DeM3FD datasets. For fair comparison, source images undergo degradation-specific preprocessing: Zero-DCE++ \cite{li2021zerodce++} for low-light, ASCNet \cite{yuan2025ascnet} for stripe noise, and Swin-IR \cite{liang2021swinir} for Gaussian noise.

\subsubsection{Qualitative Analysis}
Fig.\ref{DegenerateQualitative} presents qualitative comparisons under challenging scenarios. GANMcC and RFN-Nest suffer from imbalanced intensity inheritance, failing to preserve contrast under poor illumination (green boxes in rows 1 and 3). CUFD and DATFuse extract limited infrared information (green box, row 3), while CUFD, DATFuse, and FusionMamba struggle to retain fine details in noisy conditions (red box in row 2; green and red boxes in row 4). MDA and ITFuse produce results with insufficient contrast (red box, row 3). Moreover, two-stage methods fail to suppress residual artifacts from preprocessing, leading to noise remnants and color distortions—evident across all baselines (e.g., stripe noise in row 1, Gaussian noise in row 2, and mixed noise in rows 3 and 4). In contrast, DDFusion effectively suppresses degradations and preserves complementary information via unified degradation separation and cross-modal fusion, resulting in visually natural and perceptually superior outputs.

\subsubsection{Quantitative Analysis}
Tab.\ref{tab:Quantitative_degrade} reports quantitative results under severe degradations. Aside from a slightly lower SF on DeLLVIP compare to DATFuse, DDFusion consistently outperforms all methods, validating its effectiveness in degradation removal and complementary feature fusion by decoupling compound degradations and jointly optimizing degradation suppression and fusion processes.

\begin{table}[!t]
\centering
{\small
\begin{tabular}{ccccccc}
\toprule
 & \textbf{VIF$\uparrow$} & \textbf{AG$\uparrow$} & \textbf{EI$\uparrow$} & \textbf{Q$_{abf}$$\uparrow$} & \textbf{SF$\uparrow$} & \textbf{Q$_{w}\uparrow$}\\
\midrule
 w/o DDON& 1.0057 & 6.0064 & 61.5511 &  0.3246 & 18.8462 & 0.5671 \\
 w/o ILGFN& 0.8643 & 5.7803 & 59.2956 & 0.3169 & 17.6852 & 0.5750 \\
\midrule
 w/o $\mathcal{L}_{ds}$ & \textbf{1.0205} & 6.1119 & 62.3288 & 0.3076 & 19.3660 & 0.5482 \\
 w/o $\mathcal{L}_{text}$& 0.5402 & 2.8786 & 30.2643 & 0.2159 & 7.8931 & 0.3599 \\
 
\midrule
\textbf{$\star$DDFusion}
             & 0.9188
             & \textbf{6.3258}
             & \textbf{64.5265}
             & \textbf{0.3284}
             & \textbf{19.7924}
             & \textbf{0.5838} \\
\bottomrule
\end{tabular}
}
\caption{Quantitative comparison of ablation experiments. The best results for each metric are bolded, and $\uparrow$ indicates that higher values denote better performance.}
\label{tab:ablation}
\end{table}

\subsection{Ablation Study} \label{subsec:ablation}
\subsubsection{Qualitative Analysis:}
Fig.~\ref{Ablation} presents qualitative ablation results on the network architecture and loss functions. (1) Without DDON, the network fails to suppress source degradations, resulting in low contrast (green box, first row) and inherited infrared noise (red boxes, first and second rows). (2) Without ILGFN, edge details are poorly extracted and fused, as seen in the tire (red box, second row) and shadow (green box, first row). (3) Without $\mathcal{L}_{ds}$, DDON underperforms in degradation removal (striped noise, red box, first row) and struggles to preserve source details and intensity (red and green boxes, second row). (4) Without $\mathcal{L}_{text}$, the fused results suffer significant edge and detail loss, compromising visual quality.
\subsubsection{Quantitative Analysis:}
Tab.\ref{tab:ablation} presents the quantitative ablation results. Removing DDON or $\mathcal{L}_{ds}$ slightly increases VIF, suggesting that more visual features—along with residual degradation—are retained from the sources. The decline in other metrics indicates insufficient degradation suppression. Excluding ILGFN causes notable drops in AG, EI, and SF, underscoring its role in fusing multi-scale details and structure. Removing $\mathcal{L}_{text}$ significantly degrades all metrics, confirming its critical contribution to the fusion process.

\begin{figure}[!t]
\centering
\includegraphics[width=0.9\columnwidth]{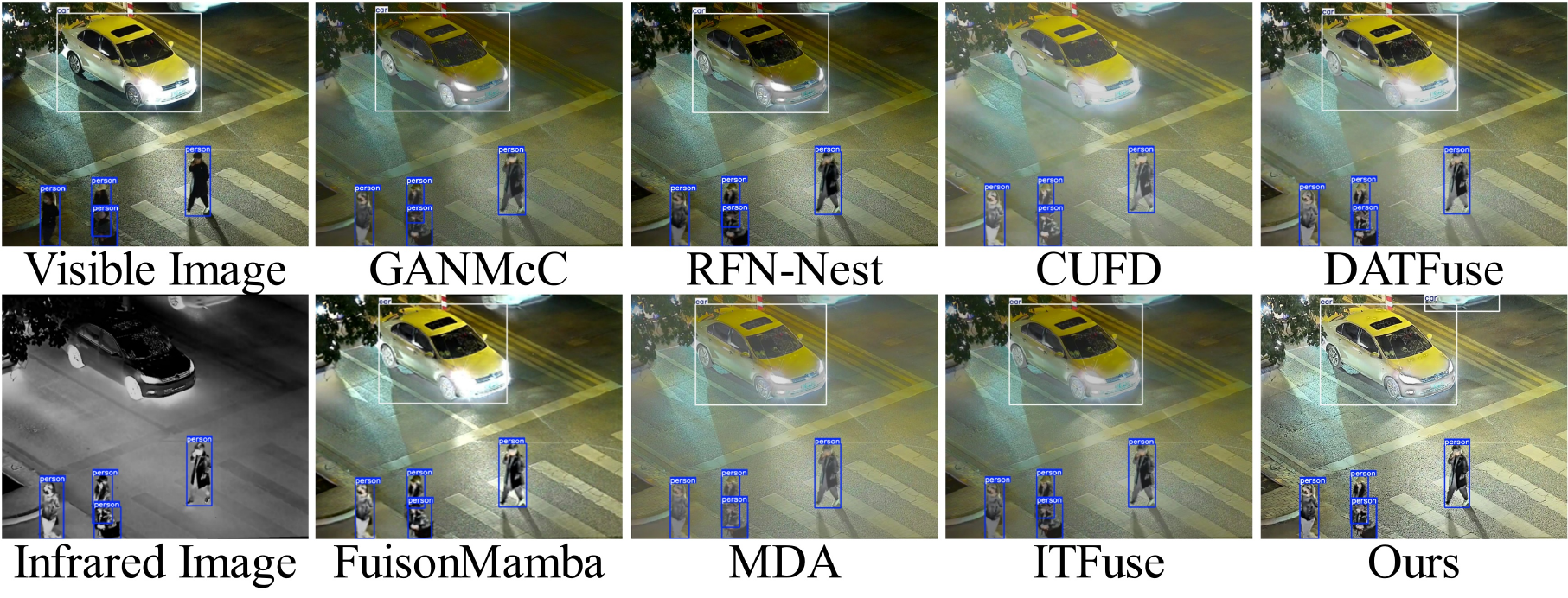}
\caption{Detection performance of DDFusion and seven comparison methods on fused image from LLVIP dataset.}
\label{Detection}
\end{figure}

\subsection{Extension Experiment} \label{subsec:visualization}
We evaluate the performance of DDFusion and competing methods on downstream object detection by applying YOLOv5 to fused images from the LLVIP dataset. As shown in Fig.\ref{Detection}, our method yields more accurate detection results, effectively identifying targets that are missed (e.g., the vehicle in the upper right) or partially detected (e.g., the pedestrian in the lower left) by other methods. This demonstrates the superior effectiveness of our fusion approach in supporting downstream tasks.

\section{Conclusion} \label{sec:conclusion}
This paper proposes a degraded image fusion framework named DDFusion, which decouples degradation and unifies degradation suppression with the fusion process, effectively addressing the challenges of existing IVIF methods in handling complex degraded inputs. The proposed Degradation-Decoupled Optimization Network(DDON) couples degradation-specific decomposition with targeted feature extraction for effective suppression and representation learning. Meanwhile, the Interactive Local-Global Fusion Network(ILGFN) aggregates complementary features across multi-scale paths and mitigates performance degradation caused by the decoupling of degradation optimization and image fusion. Extensive experiments demonstrate that DDFusion outperforms state-of-the-art IVIF methods in fusion quality.

\bibliography{aaai2026}

\end{document}